\documentclass[conference]{IEEEtran}

\IEEEoverridecommandlockouts

\pdfoutput=1
\usepackage{cite}
\usepackage{amsmath,amssymb,amsfonts}
\usepackage{algorithmic}
\usepackage{float}
\usepackage{graphicx}
\usepackage{enumitem}
\usepackage{caption}
\usepackage{subcaption}
\usepackage{textcomp}
\usepackage{xcolor}
\usepackage{hyperref}
\usepackage{booktabs,siunitx,tabularx, stfloats}
\usepackage{threeparttable}
\usepackage{svg}
\usepackage{diagbox}
\usepackage{fancyhdr}
\newcommand{\orcid}[1]{\href{https://orcid.org/#1}{\includegraphics[width=10pt]{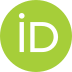}}}

\renewcommand{\baselinestretch}{0.9074}

\def\BibTeX{{\rm B\kern-.05em{\sc i\kern-.025em b}\kern-.08em
    T\kern-.1667em\lower.7ex\hbox{E}\kern-.125emX}}
\newcommand{\printfnsymbol}[1]{%
  \textsuperscript{\@fnsymbol{#1}}%
}

\begin{document}

\title{Few-shot Learning for Multi-modal Social Media Event Filtering
}

\author{José Nascimento\textsuperscript{$\star$}\orcid{0000-0003-3450-6029}, João Phillipe Cardenuto\textsuperscript{$\star$}{\orcid{0000-0002-8370-6329}}, Jing Yang\textsuperscript{$\star$}\orcid{0000-0002-0035-3960}, and Anderson Rocha\orcid{0000-0002-4236-8212}\\
Artificial Intelligence Lab. (\url{Recod.ai}), Institute of Computing, University of Campinas, SP, Brazil \\
\{jose.nascimento, phillipe.cardenuto, jing.yang\}@ic.unicamp.br, arrocha@unicamp.br\\}
% Need to insert emails and ORCID

\maketitle
\begingroup\renewcommand\thefootnote{$\star$}
\footnotetext{Equal contribution}
\endgroup

\fancypagestyle{firstpage}{
    \renewcommand{\headrulewidth}{0pt}%
    \fancyhf{}
    \fancyfoot[L]{\footnotesize\textcopyright 2022 IEEE.  Personal use of this material is permitted. Permission from IEEE must be obtained for all other uses, in any current or future media, including reprinting/republishing this material for advertising or promotional purposes, creating new collective works, for resale or redistribution to servers or lists, or reuse of any copyrighted component of this work in other works.}
 }
 
\thispagestyle{firstpage}

\begin{abstract}
Social media has become an important data source for event analysis. When collecting this type of data, most contain no useful information to a target event. Thus, it is essential to filter out those noisy data at the earliest opportunity for a human expert to perform further inspection. Most existing solutions for event filtering rely on fully supervised methods for training. However, in many real-world scenarios, having access to large number of labeled samples is not possible. To deal with a few labeled sample training problem for event filtering, we propose a graph-based few-shot learning pipeline. We also release the Brazilian Protest Dataset to test our method. To the best of our knowledge, this dataset is the first of its kind in event filtering that focuses on protests in multi-modal social media data, with most of the text in Portuguese. Our experimental results show that our proposed pipeline has comparable performance with only a few labeled samples (60) compared with a fully labeled dataset (3100). To facilitate the research community, we make our dataset and code available at \url{https://github.com/jdnascim/7Set-AL}. 
\end{abstract}

\begin{IEEEkeywords}
event filtering, few-shot learning, social media dataset
\end{IEEEkeywords}

\section{Introduction}

Since the last decade, social networks have been playing a significant role in different protests around the world. 
Because of that, an analyst that aims to better understand a given event might need to take into account social media data, that in most scenarios are images, videos, and texts. As an illustration, this analyst might be a social sciences researcher conducting a study; a journalist that works with fact-checking covering the event in real-time; or a forensic analyst in a scenario where some aspects of the event might be inspected on trial. 
However, one problem that arises when collecting data from social media for event analysis is the presence of massive irrelevant data in the retrieved result. Due to the enormous total size of the dataset, manually labeling every post as relevant might be unfeasible in a real-world scenario. 

\begin{figure*}[h]
\centering
\includegraphics[width=0.95\textwidth]{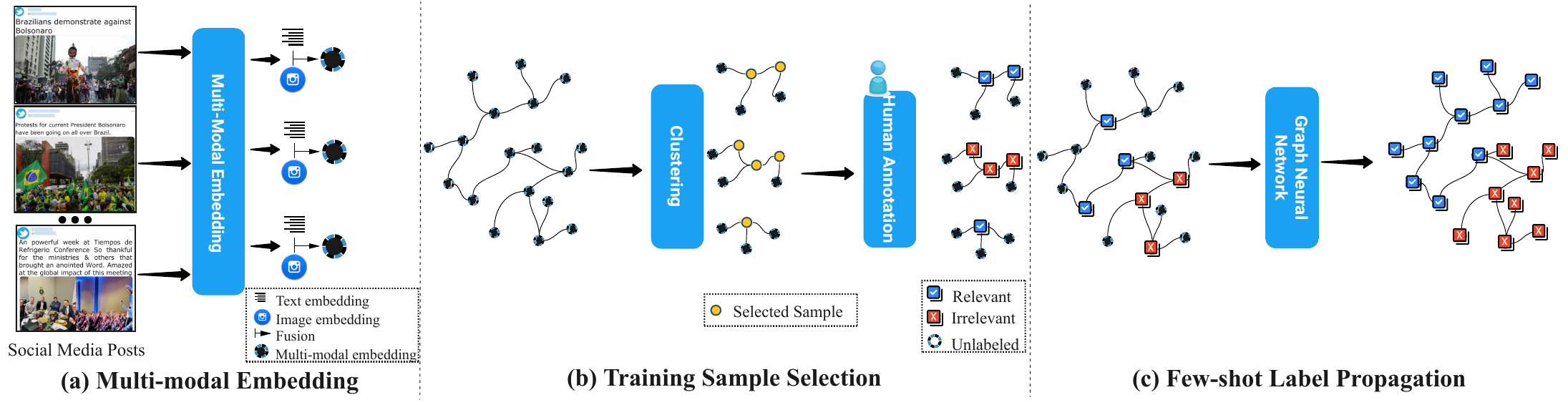}
         \caption{Our proposed few-shot event filtering pipeline. The pipeline consists of three main steps: (a) multi-modal feature embedding, (b) training sample selection and (c) few-shot label propagation. In the first step, we extract the representation of a tweet that contains two modalities: text and image; then we select key samples to be labeled based on a clustering and ranking approach; finally, we construct a graph and train a GNN-based model with few-shot learning to propagate the label to the entire graph.}
\label{fig:abstract-fig}
\end{figure*}

Recent approaches~\cite{multimodalbaseline2020, kumar2020deep, mouzannar2018damage} for event filtering employ deep learning techniques to reduce the information overload. These methods usually require thousands of labeled data for training. However,
labeling a large dataset might be unfeasible for the time and resources needed in some cases.
% how we address
To address this problem, we study the employment of deep learning methods based on few-shot learning. In detail, we propose a novel filtering pipeline, shown in Figure \ref{fig:abstract-fig}. The pipeline is composed of three main steps: multi-modal embedding, training sample selection and few-shot label propagation.

To test our method, we collected a dataset containing 4.5 million tweet posts related mainly to two large political demonstrations in Brazil on the 7th of September and on the 2nd of October, 2021. All these demonstrations depict the political polarization of Brazil, in which supporters of President Bolsonaro (President of Brazil at the time of the event) manifested on the 7th of September, and the opposition manifested on the 2nd of October. Apart from these two demonstrations, some minor protests without a clear political flag occurring on the 12th of September were also included in the database. Despite carefully querying keywords related to these events, the dataset has a large number of irrelevant and duplicated tweets, as expected in a real-world scenario. Therefore, these data cannot be directly delivered to an expert to analyze the event. In this context, our proposed pipeline aims to filter out such data with few-shot learning.

To evaluate our approach quantitatively, we selected five thousand samples (without near-duplicates) and labeled them as relevant or irrelevant with respect to the event of \textit{Brazilian Political Protests}. We investigate our few-shot learning pipeline by analyzing the three main components: multi-modal embedding models, few-shot learning models and training selection methods. The experiments show that our graph-based few-shot learning performs well with only a few carefully selected samples.

We summarize our contributions as follows:
\begin{enumerate}
    \item We propose a data-centric multi-modal few-shot learning method for event-filtering.
    \item We collect and label a multi-modal Brazilian Political Protest Dataset\footnote{\url{https://zenodo.org/record/5676056\#.Yq8zkXbMJLc}}.
    \item We provide an open-source tool for crawling tweets, related media (images, audios and videos) and news-articles\footnote{\url{https://github.com/phillipecardenuto/fakenews-crawler}}.
\end{enumerate}

\section{Proposed Event Filtering Pipeline}

To deal with the lack of annotated items in a large dataset, we propose a graph-based few-shot learning pipeline. We show the pipeline in Figure \ref{fig:abstract-fig}. The pipeline consists of three main steps: (a) multi-modal feature embedding, (b) training sample selection, and (c) few-shot label propagation. First, we extract the representation of a tweet that contains two modalities: text and image; then, we select training samples based on a clustering and ranking approach; finally, we construct a graph based on the similarity of each sample's multi-modal representation from the database and utilize a graph-based few-shot learning to label the entire graph. 

\subsection{\textbf{Multi-modal Embedding}}
A crucial step of our pipeline is creating a robust representation of a tweet post (text and related image).
This representation should provide semantic aspects for event filtering (ideally using all related post content, i.e., text and multi-media).
For this task, we use CLIP \cite{clip2021}. CLIP was trained to match an image with a text snippet using a contrastive approach \cite{clip2021}.
Due to the CLIP training process, this model can extract visual semantic features from an image and associate them with their textual description.
Also, CLIP is able to represent in the same feature-space both modalities, facilitating any multi-modal operation.

The image feature is extracted using CLIP's image encoder based on Visual Transformers (ViT) pre-trained on WebImageText (WIT) --- a dataset proposed by the CLIP authors. Because our dataset consists of multilingual tweets, with the majority of them in Portuguese, the text embeddings were extracted using a multilingual CLIP's text encoder from \textit{sentence-transformer}~\cite{sentence-bert} --- pre-trained on Portuguese, English, and Spanish, among other languages.

\subsection{\textbf{Training Sample Selection}} ~\label{sec:train-sample-selection}
Selecting which samples to train is an important factor on a model's performance~\cite{ng2021mlops}. In our pipeline, we select samples by clustering and selecting the most representative nodes from each cluster. The benefit of clustering is that it can group similar samples together, which would ideally separate relevant samples from irrelevant ones, and selecting from different clusters can bring diverse and balanced training samples. 

Specifically, we first calculate the cosine similarity matrix of the tweets' CLIP embeddings, and construct a $\epsilon$-graph with $\epsilon=0.85$~\cite{Luxburg2014}. Then, we perform Leiden community detection~\cite{traag2019louvain}, which is a graph-based clustering method that finds an optimal community partition of a graph by optimizing the modularity value of the graph.

After performing clustering, we pick the representative samples from each resultant cluster. For that, we rank the nodes within each detected community based on a centrality measure. Then, we choose the representative samples based on their ranking score within their cluster. Finally, we acquire the labels of selected items from a human expert\footnote{To simplify the procedure, here we use our annotation as the oracle label, but in real scenarios, an expert would label the selected samples.}.

\subsection{\textbf{Few-shot Label Propagation}}~\label{sec:few-shot-prop}
Given that labeling a large amount of data is unfeasible for most event cases, the number of representative samples sent to be annotated on the previous step should be small in a real scenario. Thus, we focus on semi-supervised techniques for event filtering.

In this sense, we use a Graph Neural Network (GNN) for this task. 
Differently from traditional graph-based semi-supervised learning methods (e.g., LGC~\cite{NIPS2003_87682805}) that spread the labels across a graph, GNNs can update the embeddings of the set of nodes based on the structure of the graph, by incorporating information from the respective neighbors. 

To define the neighborhood of the nodes, we choose the most similar elements to a vertex using KNN (K-Nearest Neighbors). We build a graph using the entire train set and divide the representative samples into training and validation set (keeping a balance among the labels). During training, only the selected samples are used for the GNN model loss calculation. Then, we perform training for a pre-determined number of epochs. After that, we select the model with the best accuracy on the validation set.

For inference, we construct another graph with the whole dataset (train + test). Then, we predict the labels using the best model and report the balanced accuracy on the test set. Note that, during inference, the training data is used in the final graph by improving the neighborhood of the test set and incorporating more information into the GNN, improving the label propagation process.

\section{Dataset Collection and Annotation}
To test our methods, we collect an event filtering dataset and annotate a subset for quantitive evaluation. 
Specifically, we introduce the content of our protest dataset, the application that we developed to obtain these data, and the annotation process for the five thousand samples selected from the dataset.

\subsection{\textbf{Brazilian Political Protest Dataset}}

\begin{figure}
\centering
\includegraphics[width=0.45\textwidth]{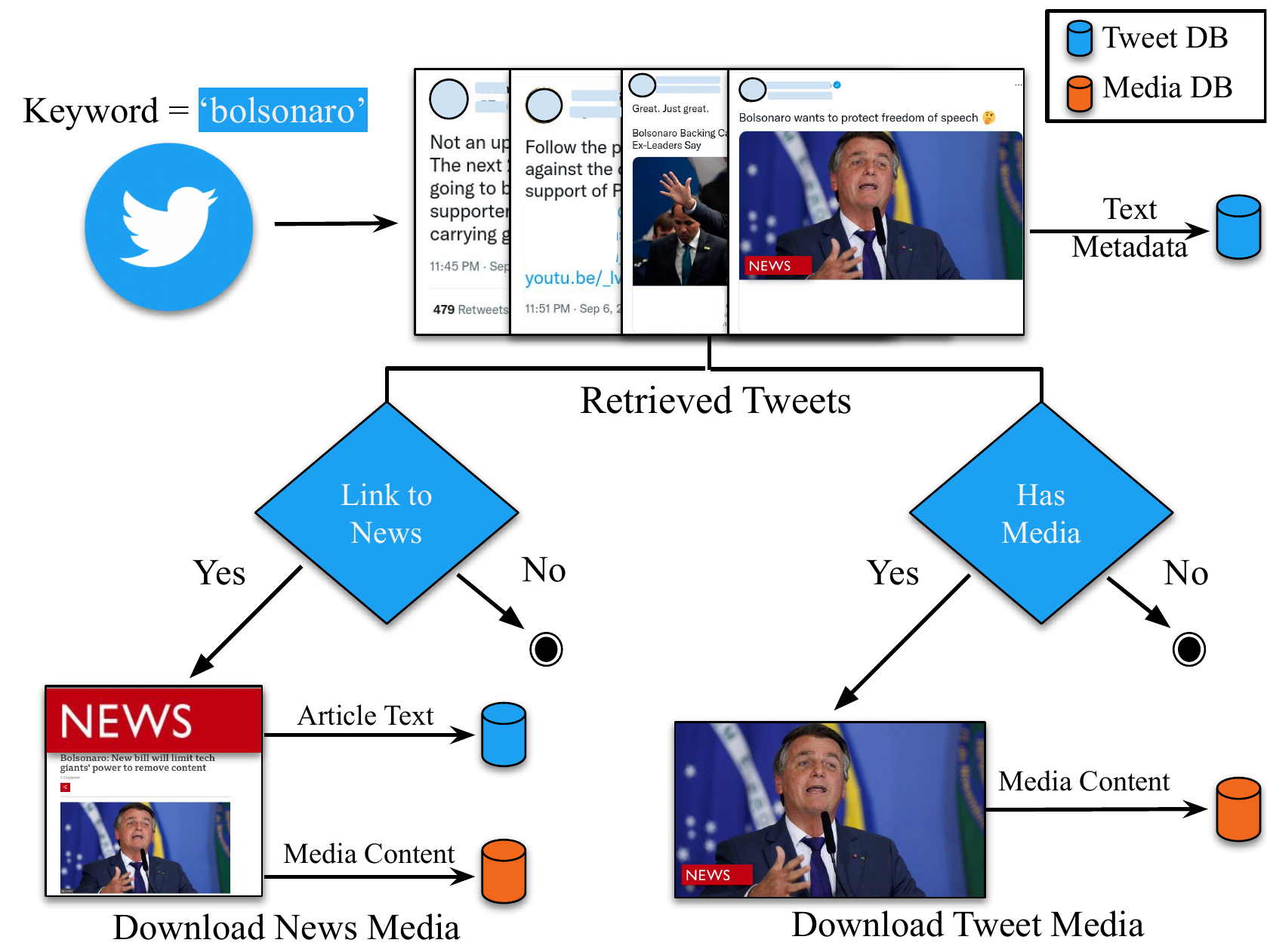}
         \caption{Tweet gathering process. We start the gathering by querying a keyword related to the event. Then, for each retrieved tweet, we store its text and metadata (e.g., date, language, tweet ID) in the Tweet Database (in blue). If the tweet has a media associated with it (e.g., an image), we store this media in a Media database (in orange). If the tweet has a link that redirects to a news article, we download the text and media from it, storing the text in the Tweet database and the media in the Media database.}
\label{fig:tweet-collection}
\end{figure}

% Dataset motivation/background
Social media has a significant impact to Brazilian political polarization. One of the platforms used for this purpose is Twitter, which presents a virtual battlefield between the two most popular politicians in Brazil (Mr. Bolsonaro and Mr. Lula). Because of this polarization, misinformation has been shared to defend or attack each other. To assist journalists in better understanding this context, we downloaded tweet posts and news articles related to two large demonstrations in Brazil in 2021. The first was on the 7th of September, when most people supported Mr. Bolsonaro’s administration. The second was on the 2nd of October when the protesters were primarily activists of left-wing parties against Mr. Bolsonaro and supporters of Mr. Lula. Both protests were held in several cities in Brazil.
During this section, we detail (a) the dataset gathering tool, (b) the collection protocol, and (c) the collected dataset.

%% Gathering Tool
\subsubsection{\textbf{Dataset Gathering Tool}} To collect the dataset, we implement an open-source tool for automatically downloading the tweets and related news articles based on keywords included in the tweets. For example, the keyword `bolsonaro' would retrieve the post ``\textit{\#Brazil \#Bolsonaro: Thousands protest calling for president's removal, check myfakeblog.com/brazilProtest}'' since it contains the word `bolsonaro'. Also, the tool would visit \textit{myfakeblog.com/brazilProtest} and download the content, given that it was an URL in the post.

As depicted in Figure \ref{fig:tweet-collection}, our tool starts querying tweets based on input keywords. Then, for each retrieved tweet, we store the text and metadata (e.g., post date, language, URL) on a document-oriented database (Tweet DB). If a retrieved tweet has an image on its post, the tool will store the media in a dedicated database (Media DB). In addition, if the tweet is linked to a news article website, the tool visits this site and downloads its article text and media, storing them in the Tweet DB and Media DB, respectively. We released our tool at \url{https://github.com/phillipecardenuto/fakenews-crawler}.

% Gathering Protocol 
\subsubsection{\textbf{Dataset Collection Protocol}} We empirically select the groups of keywords that retrieved the most related content to the event. Table \ref{tab:group-keywords} presents the selected keyword groups.
Part of the empirical process consists of checking the Brazilians' Twitter trending topics and news headlines related to the event. Each group regards the keywords that simultaneously appeared in the trending topics. To avoid missing tweet posts suspected of misinformation spreading --given the Twitter policy of deleting such tweets--, we start gathering the data while the demonstrations occurred and during their following days.

\subsubsection{\textbf{Dataset Numbers}} The dataset contains 4.5 million tweets, of which 155 thousand are associated with an URL to an uncurated article and 370 thousand have an associated media content (including the media of the uncurated articles). Table \ref{tab:dataset-summary} summarized the collected data, presenting for each group of keywords (from Table \ref{tab:group-keywords}) the collected period and the number of tweets, media, and news articles retrieved.

\begin{table*}[t!]
\centering
\caption{Group of Keywords used for data collection.}
\footnotesize
\begin{tabular}{ll}
\toprule
Group & Keywords                                                                                                                                     \\
\midrule
G-1   & `7deSetPelaLiberdade', `ImpeachmentBolsonaroUrgente', `VoltaLula', `Venceremos', `AnonymousBrof', `SeteDeSetembro', `12deSet' \\
G-2   & `SupremoÉOPovo', `Lula', `bozo', `bolsonaro', 'bolsomito', `bolsominion', `bolsonarista', `flopou'                                          \\
G-3   & `psdb', `novo', `vemprarua', `EleNao', `BolsonaroAte2026',`BolsonaroOrgulhoDoBrasil', `Dia07EuVou'                                          \\
G-4   & `manifestação', `protesto', `Impeachment', `Dia12PeloImpeachment', `BolsonaroAcabouOR', `ForaBolsonaroGenocida', `ForaCorno'                \\
G-5   & `comunista', `fascista', `adnet'                                                                                                            \\
G-6   & `EstadoDeSitio', `Caminhoneiros', `ForaBolsonaro',`DomingoForaBolsonaro'                                                                  \\
G-7   & `DomingoEmCasaComLula', `12SetEuNaoVou', `12SetEuVou',`MBL', `DomingoAdeusBolsonaro', `NemBolsonaroNemMBL'                               \\
G-8   & `CiroNasRuas', `aculpaédobolsonaro', `VoltaLula', `ImpeachmentJá', `2outeuvou', `ForaBolsonaro', `2OutForaBolsonaro', `lula2022'    \\
\bottomrule
\end{tabular}
\label{tab:group-keywords}
\end{table*}

\begin{table}[h!]
\centering
\caption{Brazilian Political Protest Dataset Content.}
\footnotesize
\smallskip 
\begin{tabular}{llrrr}
\toprule
\multicolumn{1}{l}{Keyword Group} & \multicolumn{1}{l}{Period} & \multicolumn{1}{l}{\#Tweets} & \multicolumn{1}{l}{\#Media} & \multicolumn{1}{l}{\#News Articles} \\
\midrule
G-1                               & 6-13 Sept                            & 1.5M                              & 0.14M                    & 80K                             \\
G-2                               & 7-13 Sept                       & 1.3M                             & 96K                       & 27K                             \\
G-2                               & 1-3 Oct                          & 0.15M                               & 20K                    & 11K                             \\
G-3                               & 7-13 Sept                       & 0.76M                               & 55K                     & 27K                             \\
G-4                               & 6-13 Sept                       & 0.2M                               & 18K                     & 2K                              \\
G-5                               & 7-13 Sept                       & 0.18M                               & 12K                     & 974                               \\
G-6                               & 7-13 Sept                       & 0.16M                               & 11K                     & 520                               \\
G-7                               & 11-13 Sept                      & 89K                                & 8K                      & 3K                                 \\
G-8                               & 1-3 Oct                          & 2.6K                                 & 500                       & 47                               \\
\bottomrule
\end{tabular}
\label{tab:dataset-summary}
\end{table}

\subsection{\textbf{Event Filtering Dataset}}
To evaluate our approach for event filtering, we annotate five thousand unique samples of the \textit{Brazilian Political Protest
Dataset}. 
For that, we start randomly selecting a hundred thousand tweet posts with at least one image related to it. Given that re-tweets and near-duplicated posts are frequent on the Twitter platform, we remove tweets in which the image or text was near-duplicated for a more representative sampling.

After starting the annotation process, we notice that political protest posts have semantic ambiguity. While a tweet image of a crowded place with people hanging signs in protest against Mr. Bolsonaro administration is clearly a relevant post, a meme of Mr. Bolsonaro referring to the demonstration might not present a straightforward relationship with the event. Aiming to address this fuzziness, we define four categories to annotate text and images from a tweet:

\begin{itemize}%[{leftmargin=3.5mm}]
    \item \textbf{Related Informative}: The content from the text/image clearly presents elements related to the demonstration.
    \item \textbf{Related}: The content from the text/image does not present a clear element regarding the political demonstration, but due to the political context, it is related to the demonstration.
    \item \textbf{Irrelevant}: The content from the text/image is not related to the demonstration.
    \item \textbf{Not Sure}: The annotator was not able to decided whether the content is \textit{Related} or \textit{Irrelevant}.
\end{itemize}

Three of the authors annotated the five thousand selected tweets using the criteria above, with each modality of the tweet (text and image) annotated separately. Thus, a tweet could have a text classified as \textit{Irrelevant} but its images as \textit{Related Informative}.

We translate the labels from each modality into 2 classes, \textit{Relevant} or \textit{Irrelevant} to the event. For that, we consider an image/text as \textit{Relevant} if at least 2 annotators considered it as \textit{Related Informative} or \textit{Related}, otherwise it is considered \textit{Irrelevant}.
The annotators unanimously agreed on both modalities in $89\%$ of the cases.

Given that each modality of the tweet was labeled separately, at the end of the annotation process, we have three types of relevance: Image, Text, and Tweet.
Image relevance expresses the agreed label of the tweet's image; Text relevance, the agreed label of the tweet's text; and Tweet relevance convey whether the tweet image or text is \textit{Relevant}.
Table \ref{tab:dataset-distribution} shows the labels distribution for each modality of the dataset. We split the data into 3100 samples for training and 1900 for~testing.

\begin{table}[]
\centering
\caption{Event Filtering Dataset Sample Distribution}
\centering
\begin{threeparttable}
\centering
\begin{tabular*}{0.8\columnwidth}{lrrrr}
\toprule
         & \multicolumn{2}{c}{Train (3100)} & \multicolumn{2}{c}{Test (1900)} \\
Modality & Relevant        & Irrelevant        & Relevant        & Irrelevant       \\
\midrule
Image    & 23\%            & 77\%              & 16\%            & 84\%             \\
Text     & 37\%            & 63\%              & 25\%            & 75\%             \\
Tweet\tnote{*}     & 39\%            & 61\%              & 26\%            & 74\%    \\        
\bottomrule
\end{tabular*}
\begin{tablenotes}[flushleft]
  \item[*]A tweet is \textit{Relevant} if its text or image is \textit{Relevant}.
  \end{tablenotes}
\end{threeparttable}
\label{tab:dataset-distribution}
\end{table}

\section{Experiments and Analysis}

In this section, we report our experiments aiming at understanding three aspects related to few-shot learning pipeline for event filtering: (\ref{subsec:embedding}) Data Representation, (\ref{subsec:model}) Few-Shot Modeling, and (\ref{subsec:selection}) Sample Selection.

\subsection{\textbf{Multi-model Media Data Embeddings}}
\label{subsec:embedding}
    
To understand which social media post representation (text, image, or both) is more suitable for event filtering, we tested off-the-shelf deep learning models for text and image feature extraction. The chosen models were \textit{MobileNetV3}\cite{mobilenetV3} pre-trained on ImageNet; \textit{RoBERTa}\cite{Liu2019RoBERTaAR} pre-trained on a Tweet text database\cite{barbieri2020tweeteval}; and \textit{CLIP}\cite{clip2021} fine-tuned in a multilingual text dataset, which the weights\footnote{\url{https://huggingface.co/sentence-transformers/clip-ViT-B-32-multilingual-v1}} were provided by \textit{sentence-transformer}~\cite{sentence-bert}. 

Our evaluation considers the text embeddings extracted using \textit{RoBERTa} and CLIP; image embeddings extracted using \textit{MobilNetV3} and \textit{CLIP}; the concatenation ($\oplus$) of \textit{MobileNetV3} and \textit{RoBERTa} embeddings;
the concatenation ($\oplus$) of \textit{CLIP} image and text embeddings; and the element-wise addition ($+$) of \textit{CLIP} image and text embeddings.

We tested the performance of each evaluated feature by training a multi-layer perceptron classifier (MLPC) fed with these features. During training, we used all 3100 tweets from the train set. Also, we use the same MLPC architecture during all experiments, that has three hidden layers of size 128, 128, and 64, followed by the classifier. Given that the MLPC's architecture was the same in all experiments, we expect that the performance of the model is deeply linked to the quality of each evaluated embedding.

We evaluated the descriptors' quality considering three relevance modalities: \textit{image}, \textit{text}, and the entire \textit{tweet}. In the \textit{image} modality, only the tweet's image relevance is considered (i.e., a tweet will be considered \textit{Relevant} if its image was annotated as \textit{Relevant}, regardless of the tweet text annotation). Similarly, in the \textit{text} scenario, only the tweet text relevance is considered. In the \textit{tweet} modality, the post was considered \textit{Relevant} if any of its text or image content was labeled as \textit{Relevant}.

Table \ref{tab:emb-exp} presents the results of our evaluation in the test set using the balanced accuracy metric. The table indicates that the fusion of text and image embeddings improves the representation of a tweet. For instance, \textit{CLIP Img$\oplus$Text} embedding performs better than \textit{CLIP} single modal in most scenarios for predicting the relevance of tweets. It also shows that the \textit{CLIP} features have a better representation for social media event filtering than the others, providing $87.5\%$ balanced accuracy in comparison to $83.9\%$ of \textit{RoBERTa$\oplus$MobileNetV3} in the \textit{tweet} relevance modality.

\begin{table*}[]
\caption{Embedding performance training a MPLC with different relevance label modalities.}
\begin{threeparttable}
\begin{tabular}{lccccccc}
\toprule
\diagbox[width=\dimexpr \textwidth/8+2\tabcolsep\relax, height=1cm]{Label Modality}{ Embedding }
 & \begin{tabular}[c]{@{}c@{}}\textit{RoBERTa}\\(Bacc\%)\end{tabular} & \begin{tabular}[c]{@{}c@{}}\textit{CLIP Text}\\(Bacc\%)\end{tabular}  & \begin{tabular}[c]{@{}c@{}}\textit{MobileNetV3}\\(Bacc\%)\end{tabular}  & \begin{tabular}[c]{@{}c@{}}\textit{CLIP Image}\\(Bacc\%)\end{tabular} &
 \begin{tabular}[c]{@{}c@{}}\textit{MobileNetV3$\oplus$RoBERTa}\\(Bacc\%)\end{tabular}  &
 \begin{tabular}[c]{@{}c@{}}\textit{CLIP Image$\oplus$Text}\\(Bacc\%)\end{tabular} &
 \begin{tabular}[c]{@{}c@{}}\textit{CLIP Image+Text}\\(Bacc\%)\end{tabular}  \\
 \midrule
Text                                                               & 83.1    & 84.7      & 71.7          & 80.0          & 84.4                       & \textbf{86.3}          & \textit{86.1}     \\
Image                                                              & 65.7    & 69.1      & 77.1 & \textbf{86.2} & 77.6                       & \textit{86.0}          & 78.9              \\
Tweet (Image and Text)    & 82.3    & 84.9      & 73.8          & 81.8          & 83.9                       & \textbf{87.5}          & \textit{87.0}    \\
\bottomrule 
\end{tabular}
\begin{tablenotes}[flushleft]
  \item $\oplus$: Concatenation, +: Element-wise addition;
  \item Best in \textbf{bold}, second in \textit{italic};
  \item Bacc\%: Balanced accuracy in percentage.
D
  \end{tablenotes}
\end{threeparttable}
\label{tab:emb-exp}
\end{table*}

\subsection{\textbf{Few-shot Learning Models}}
\label{subsec:model}
To study if our graph-based learning method is suitable for few-shot event filtering, we compare our GNN models with two different types of layers in addition to two other baselines, that are listed below:

\begin{itemize}
    \item \textbf{NGCNLin (Ours)}: As we mentioned in section~\ref{sec:few-shot-prop}, we targeted to experiment GNNs in this task. The first model is a Graph Convolutional Network (GCN) layer~\cite{DBLP:conf/iclr/KipfW17} followed by a linear layer for classification. We setup the hidden size to 4096 and trained for 1000 epochs with learning rate of $1\text{e-}5$ and weight decay of $1\text{e-}3$. For training, we constructed a KNN graph with $k=10$ using the training set, and for testing we use the same method, but with $k=16$.
    
    \item \textbf{NSAGELin (Ours)}: We experimented a variation of the previous model by exchanging the GCN layer by a GraphSAGE layer~\cite{NIPS2017_5dd9db5e} (we set the aggregation method of the layer to be ``mean''). For the other hyper-parameters, we kept the same values as NGCNLin.
    
    \item \textbf{LGC}: As a baseline, we employed a traditional semi-supervised algorithm to compare with the GNN approach, namely the Local and Global Consistency algorithm (LGC)~\cite{NIPS2003_87682805}. The idea is to propagate the labels from the annotated samples to their closest neighbors through a graph using the collective structure of the items in the embedding space. We build a KNN graph ($k=16$) for this case.  
    
    \item \textbf{MLPC}: We also compare the GNNs with the MLPC model from the section \ref{subsec:embedding}.

\end{itemize}
%%%%%%%%%%%%%%

\begin{figure}[!t]
\centering
\includegraphics[width=\columnwidth]{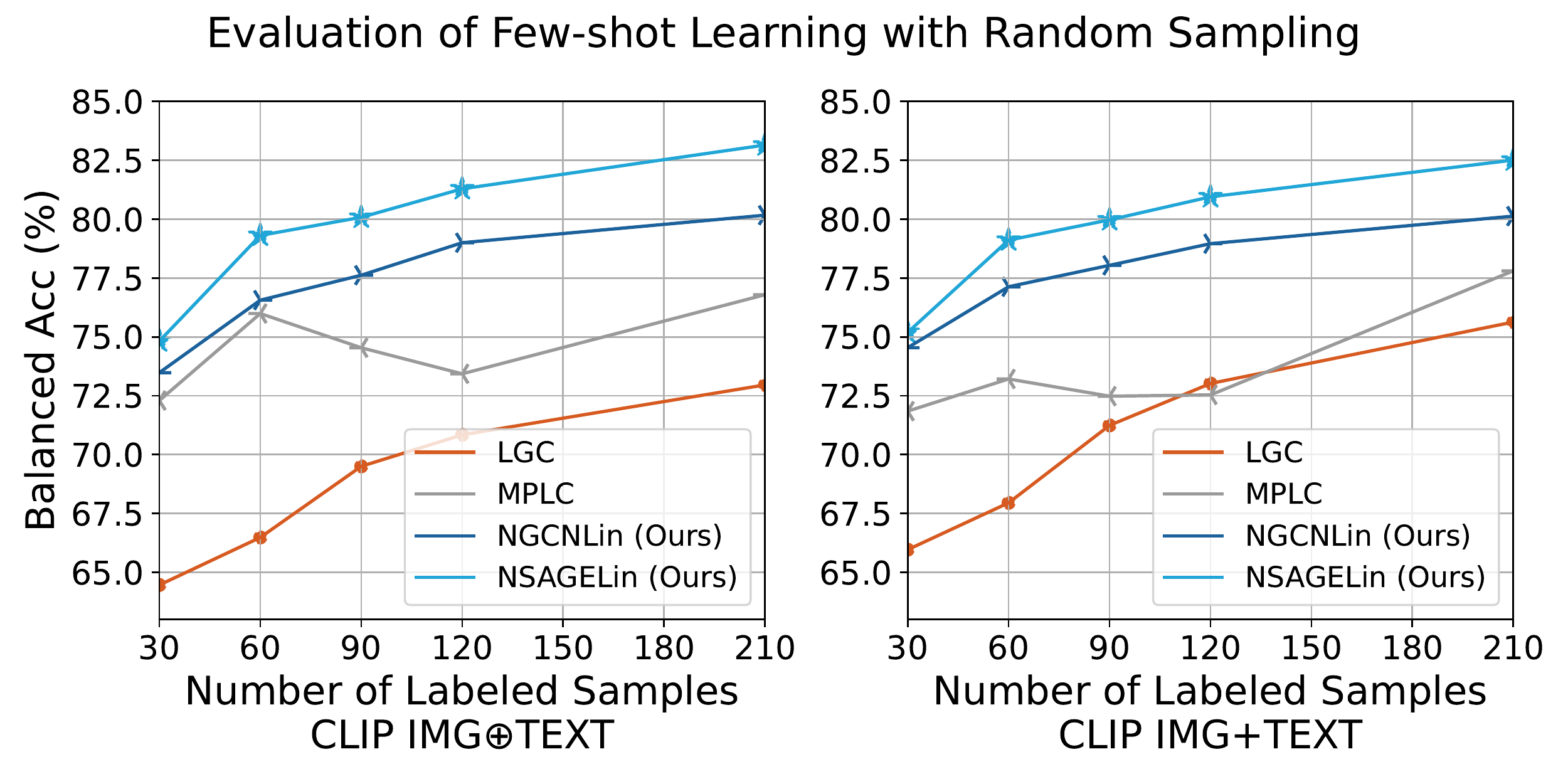}
         \caption{Evaluation of few-shot learning models with respect to different numbers of labeled data using random sampling. Each colored trace refers to a different semi-supervised model. The two best methods are NGCNLin and NSAGELin,  proposed in this work.} 
\label{fig:r2}
\end{figure}

To investigate the performance of the four models for few-shot learning, we randomly selected a number of samples from the train set ranging from 30 to 210. For each model and each number of samples, we run ten times with the same hyper-parameters, and reported the mean balanced accuracy on the test set.
Figure \ref{fig:r2} shows the performance of each model in the test set when trained using few labeled samples on top of the multi-modal CLIP embeddings. Our GNN model NSAGELin performs the best in all scenarios, followed by NGCNLin, demonstrating the capability of GNN-based models for few-shot learning.

One of the special characteristic of the GraphSAGE in comparison to GCN is that it concatenates the current node embedding with the aggregated neighborhood embeddings. We believe that this behaviour of GraphSAGE enhances the quality of the trained GNN model, since it can keep more information from the CLIP embeddings.

\subsection{\textbf{Training Sample Selection Methods}}
\label{subsec:selection}

To investigate the impact of training sample selection on model performance, we conducted experiments varying the number of labeled samples and representative sample selection methods. We first performed Leiden community detection to cluster the training samples, then selected samples based on four strategies listed below:

\begin{itemize}
    \item \textbf{Random}: randomly selects from the whole training set;
    \item \textbf{Betweenness}: ranks nodes based on the number of shortest paths that include those nodes;
    \item \textbf{PageRank}: ranks nodes based on the number of in-degrees and the importance of the corresponding source nodes;
    \item \textbf{MCI}~\cite{vega2019multi}: ranks nodes by combining multiple centrality measures (here we include: Degree, PageRank, Betweenness, Closeness, Eigenvector and Structural Holes, as proposed in the original paper).
\end{itemize}

\begin{figure}[!t]
\centering
\includegraphics[width=\columnwidth]{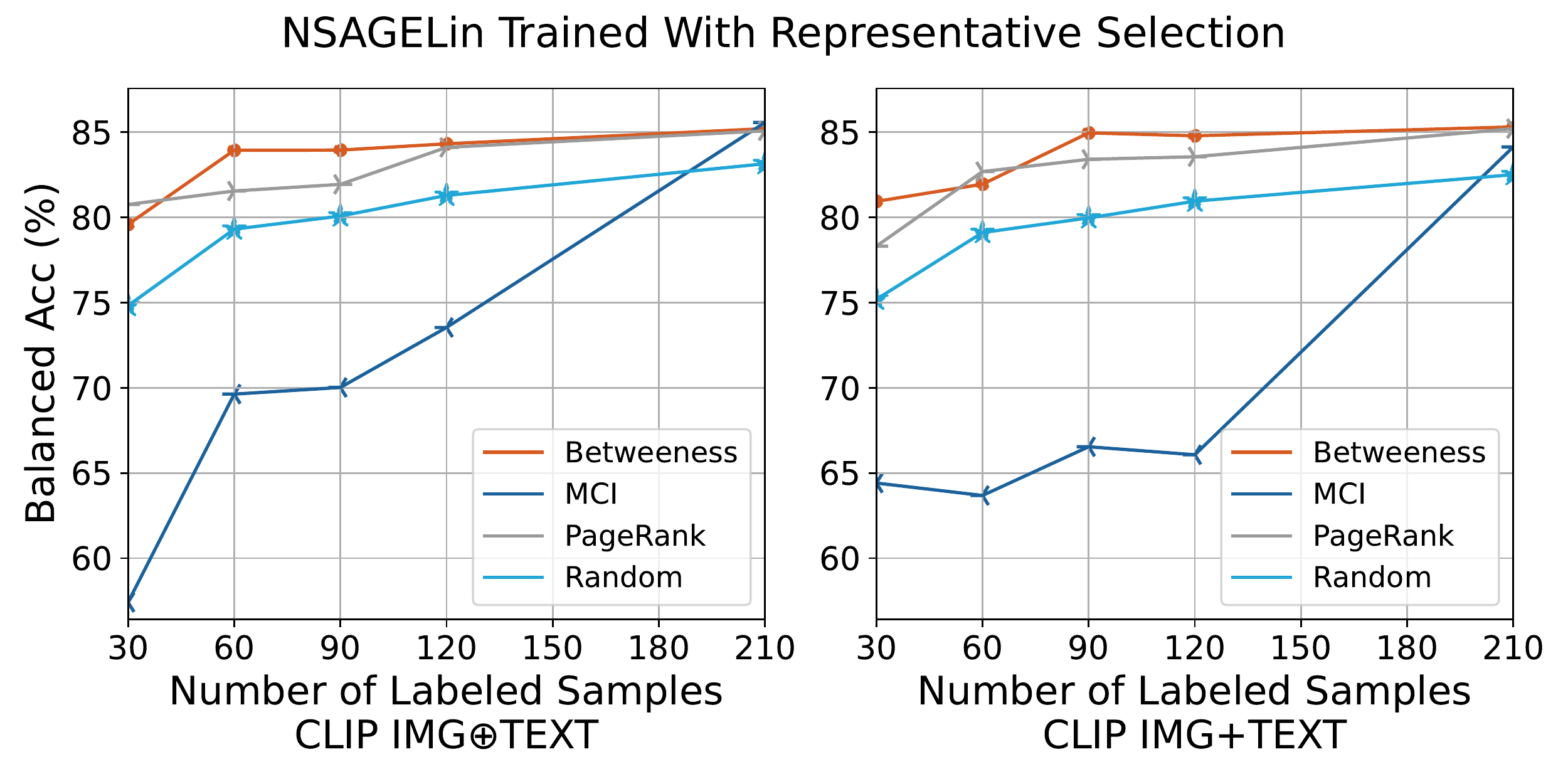}
         \caption{Classification evaluation comparison with respect to different number of samples and different sample selection methods. Each colored trace refers to a different sample selection technique. The accuracy using the full training set for concatenation ($\oplus$) and element-wise addition ($+$) are: 87.34\% and 87.28\%, respectively.}
\label{fig:r3}
\end{figure}

We tested the results with the embeddings from CLIP concatenation and addition model, respectively, followed with the few-shot learning model NSAGELin. We varied the number of labeled samples on 30, 60, 90, 120, and 210. We show the classification accuracy with respect to different number of training samples and sample selection methods in Figure \ref{fig:r3}. 

Comparing different selection methods, we notice that Betweenness has the best performance overall. Our intuition is that Betweenness provides more diverse sampling as the selected nodes serve as bridges that link different parts of the graph. On the other hand, although MCI combines multiple centrality measures, it performs the worst when the sample size is small, indicating that combining different measures is not ideal for graph-based sample selection with small sample~size.

Comparing the number of labeled samples for training, as a few-shot learning method, the fewer labeled samples, the better it is in terms of human labor. In Figure \ref{fig:r3} we can see that there is a performance gap between training with 30 samples (best around 80\%) and the entire 3100 training set (around 87\%). Still, the performance improved drastically just with 60 samples and then increased slowly with more than that (except for MCI). This indicates the capability of our graph-based label propagation method to learn with a very small labeled sample size.

% \end{enumerate}

\section{Related Work and Discussion}

Despite the area of event analysis contains various types of datasets (e.g., fire, flood, earthquake~\cite{2021arXiv210812828A}), we found only one image dataset related to protests~\cite{won2017protest}. This dataset includes ten thousand images from different protests around the world, and provides labels for different visual features. However, it lacks textual data and other metadata. Therefore, one of our contributions is the collection of a multi-modal protest dataset and annotation of a subset of these data. Due to the location and context of the target event, most items from our dataset are in Portuguese, with a small fraction in English and Spanish.

Concerning event filtering for multi-modal social media data, most work relies on separate images and text feature extraction models. Then, these features are joined into a fused representation to perform the classification. 
This joint representation is usually obtained by training a fully-connected neural network~\cite{multimodalbaseline2020, kumar2020deep, mouzannar2018damage}. The problem with these works is that they do not consider the few-shot scenario, where constructing a good model based supervised learning techniques can be hard~\cite{van2020survey}.

Although some works focus on few labels for single modality event filtering 
using active learning~\cite{said2021active} or semi-supervised learning~\cite{9414461}, there is no work focused on few-shot learning with multi-modal event data, to the best of our knowledge. In this context, one advantage of our work is presenting a new data-centric perspective to this task, by using a pipeline that trains with small (but curated) set.
We demonstrated by experiments that our sample selection method outperforms random selection by a considerable margin, and obtains results comparable with solutions trained with the full training set.

We found only one work using GNNs for multi-modal event filtering~\cite{li2022mgmp}. It proposes a pipeline that incorporates image and text features by information propagation over graphs. However, it does not considered a few-shot scenario, which we believe is necessary for real-world applications. Thus, we consider that another advantage of our pipeline is showing how GNNs can have competitive results with only a small set of labeled samples.

\section{Conclusions and Future Work}

The need for labeled data can be a bottleneck for the event filtering task.
To deal with this issue, we proposed a novel graph-based few-shot learning pipeline for multi-modal social media data. To validate our method, we collected a dataset with 4.5 million tweets related to Brazilian Political Protest in 2021 and manually labeled five thousand samples for evaluation. 
We show that large pre-trained multi-modal models (e.g., \textit{CLIP}) can extract meaningful semantics from multi-modal social media data. One important lesson from this work is that it is possible to maintain model's performance with few but carefully curated samples (data-centric approach). Our results show that when trained with few (60) but key annotated samples, GNNs can achieve comparable performance as training with a much larger annotated training set. In a real-world scenario, our method can largely reduce annotation~efforts. 

In the future, we plan to incorporate heterogenous graphs as inputs for the GNN models, with the metadata from social media posts. For example, we can input to GNNs a graph with different types of nodes (e.g., text and images) and edges representing different relations (e.g., media from the same post, similar images, similar text, etc.). 
We also plan to explore self-supervised learning techniques for event filtering --- which requires few or no labeled data ---
given the recent advances in this field for multiple tasks, including the creation of rich data descriptors for a classification task. For example, we can use the few labeled samples to retrieve similar items and utilize contrastive learning to train a better data descriptor.

 \section*{Acknowledgment}
 
This research is funded by São Paulo Research Foundation (FAPESP) under the grants DéjàVu \#2017/12646-3, \#2019/04053-8, \#2020/02241-9 and \#2020/02211-2.

\bibliographystyle{IEEEtran}
\bibliography{references}

\end{document}